\documentclass[sigconf]{acmart}

\setcopyright{none}
\renewcommand\footnotetextcopyrightpermission[1]{} 
\AtBeginDocument{%
  }

\usepackage{url}            
\usepackage{booktabs}       
\usepackage{amsfonts}       
\usepackage{nicefrac}       
\usepackage{microtype}      
\usepackage{xcolor}         
\usepackage{multirow}
\usepackage{enumitem, tabularx}
\usepackage{wrapfig,lipsum,booktabs}
\usepackage{amsmath}
\usepackage{bbding}
\usepackage{enumitem, tabularx}
\usepackage{adjustbox}

\begin{document}

\title{Saliency-Bench: A Comprehensive Benchmark for Evaluating Visual Explanations}
\author{Yifei Zhang}
\email{yifei.zhang2@emory.edu}
\affiliation{%
  \institution{Emory University}
  \city{Atlanta}
  \state{GA}
  \country{USA}
}

\author{James Song}
\email{james.song2@emory.edu}
\affiliation{%
  \institution{Emory University}
  \city{Atlanta}
  \state{GA}
  \country{USA}
}

\author{Siyi Gu}
\email{sgu33@stanford.edu}
\affiliation{%
  \institution{Stanford University}
  \city{Stanford}
  \state{CA}
  \country{USA}
}

\author{Tianxu Jiang}
\email{tianxuj@umich.edu}
\affiliation{%
  \institution{University of Michigan}
  \city{Ann Arbor}
  \state{MI}
  \country{USA}
}

\author{Bo Pan}
\email{bo.pan@emory.edu}
\affiliation{%
  \institution{Emory University}
  \city{Atlanta}
  \state{GA}
  \country{USA}
}

\author{Guangji Bai}
\email{guangji.bai@emory.edu}
\affiliation{%
  \institution{Emory University}
  \city{Atlanta}
  \state{GA}
  \country{USA}
}

\author{Liang Zhao}
\email{liang.zhao@emory.edu}
\affiliation{%
  \institution{Emory University}
  \city{Atlanta}
  \state{GA}
  \country{USA}
}

\renewcommand{\shortauthors}{Zhang et al.}

\begin{abstract}
Explainable AI (XAI) has gained significant attention for providing insights into the decision-making processes of deep learning models, particularly for image classification tasks through visual explanations visualized by saliency maps. Despite their success, challenges remain due to the lack of annotated datasets and standardized evaluation pipelines. In this paper, we introduce Saliency-Bench, a novel benchmark suite designed to evaluate visual explanations generated by saliency methods across multiple datasets. We curated, constructed, and annotated eight datasets, each covering diverse tasks such as scene classification, cancer diagnosis, object classification, and action classification, with corresponding ground-truth explanations. The benchmark includes a standardized and unified evaluation pipeline for assessing faithfulness and alignment of the visual explanation, providing a holistic visual explanation performance assessment. We benchmark these eight datasets with widely used saliency methods on different image classifier architectures to evaluate explanation quality. Additionally, we developed an easy-to-use API for automating the evaluation pipeline, from data accessing, and data loading, to result evaluation. The benchmark is available via our website:~\url{https://xaidataset.github.io}.
\end{abstract}

\maketitle
\pagestyle{plain}

\begin{figure*}[ht]
    \centering
    \includegraphics[width=\linewidth]{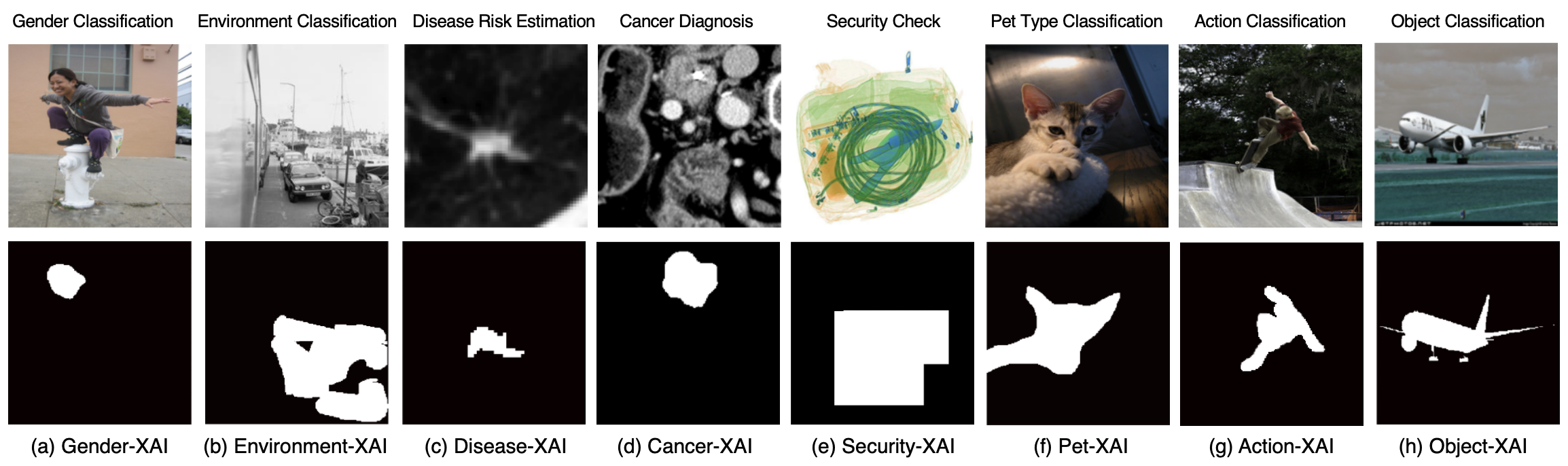}
    \caption{Example images from the eight datasets—Gender-XAI, Environment-XAI, Disease-XAI, Cancer-XAI, Security-XAI, Pet-XAI, Action-XAI, and Object-XAI—across different tasks. Each image is paired with a ground-truth explanation annotation.}
    \label{fig:teaser}
\end{figure*}

\section{Introduction}
Deep Neural Networks (DNNs) have achieved significant success in automated decision-making tasks, particularly in image classification tasks. However, their ``black box'' nature presents challenges in ensuring transparency and interpretability~\cite{adadi2018peeking, tjoa2020survey}. To address these challenges, explainable AI (XAI) techniques have emerged, providing insight into the rationale behind the model prediction process~\cite{buhrmester2021analysis, rojat2021explainable}. Among these techniques, saliency methods have gained considerable attention for their ability to generate visual explanations, enhancing user understanding and trust. By highlighting the regions most influential in model predictions, saliency maps provide valuable information about the model’s focus and rationale~\cite{zhou2018interpretable, wang2020score, ivanovs2021perturbation}. For example, when diagnosing an image as ``cancerous,'' the model’s reasoning should focus on the lesion areas, rather than on artifacts, ensuring that its prediction is based on correct reasoning.

Given the importance of saliency methods in improving model transparency, it is crucial to assess the quality of the explanations they generate. Evaluation of model explanations involves two key aspects: 1) whether the explanation reflects the true underlying reasoning of the model, and 2) how closely the explanation aligns with the ground truth. To assess these aspects, we evaluate both faithfulness and alignment~\cite{jin2023xai,doshi2017towards,nauta2023anecdotal,li2020quantitative} of visual explanations. Faithfulness measures how accurately the explanation represents the model’s true reasoning process, while alignment gauges how well the explanation corresponds to human understanding or ground truth. Together, these evaluations ensure that the model’s predictions are not only accurate but also interpretable and aligned with human understanding and the true decision-making process.

However, it is challenging to obtain explanation annotations because of prohibitive human effort and additional challenges: (1) \textit{Lack of a standardized evaluation framework}: Saliency maps are inherently continuous representations, while human annotations are usually discrete or categorical in nature~\cite{sigut2023depth, szczepankiewicz2023ground}. Such discrepancies, however, hinder meaningful comparisons between AI-generated explanations and human annotations and reduce the reproducibility of results, complicating the integration of saliency methods into real-world applications~\cite{kazmierczak2024benchmarking,bertrand2022cognitive}. (2) \textit{Absence of comprehensive and diverse datasets}: There are limited benchmark datasets focusing on limited domains, basically medical imaging and object classification~\cite{saporta2022benchmarking, mohseni2021quantitative}, which are too narrow in scope to holistically evaluate XAI. Furthermore, those relying on user interfaces~\cite{gao2022aligning, selvaraju2017grad} or questionnaires~\cite{selvaraju2017grad, petsiuk2018rise, doshi2017towards}, are expensive and not scalable. This gap in high-quality, annotated datasets significantly hampers both the development and reliable evaluation of saliency methods. (3) \textit{Lack of comprehensive benchmarks and analysis}: First, there are very few existing works, each of which, however, provides only one or a couple of datasets for evaluation. For example, \cite{gao2022aligning} introduces two datasets under medical imaging types for evaluating faithfulness but not alignment, while \cite{selvaraju2017grad} provides a low-resolution image dataset for evaluating object detection but not classification. However, the formats and evaluation metrics of these datasets are arbitrarily different from each other, making it difficult to test saliency methods across different domains and tasks.


In this work, we establish Saliency-Bench, a comprehensive benchmark for visual explanation of image classification tasks. Saliency-Bench is a collection of eight datasets with annotated ground-truth explanations, covering a wide range of topics including gender classification, environment classification, action classification, object classification, cancer diagnosis, disease estimation, pet type classification, and security check classification. These datasets are processed into a unified format, enabling consistent evaluation across different tasks. We conducted extensive benchmarking experiments to evaluate several saliency methods—GradCAM~\cite{selvaraju2017grad}, GradCAM++\cite{chattopadhay2018grad}, Integrated Gradients~\cite{sundararajan2017axiomatic}, InputXGradient~\cite{shrikumar2016not}, Occlusion~\cite{zeiler2014visualizing}, and RISE~\cite{petsiuk2018rise}—across different backbone classifier architectures including ResNet-18 and VGG-19. We also benchmark the ViT-B/16’s attention mechanism~\cite{dosovitskiy2020image} as a saliency method. For alignment evaluation, we use two key metrics: mean Intersection over Union and Pointing Game~\cite{zhang2018top}. Additionally, we assess the faithfulness of these methods using the insertion Area Under the Curve (iAUC)~\cite{petsiuk2018rise} and conduct an inter-method reliability analysis. Through these experiments, we aim to address the challenges outlined earlier by providing a standardized evaluation framework, a diverse collection of annotated datasets, and comprehensive benchmarks for saliency methods.
Our contributions are summarized as follows:
\begin{itemize}[noitemsep, leftmargin=10pt]
\item \textbf{Comprehensive Dataset Collection}: We curated, constructed, and annotated a dataset collection designed to benchmark saliency methods for image classification tasks. The collection includes eight datasets spanning topics such as action classification, tumor classification, and object classification. Each dataset comes with class labels and ground-truth explanation annotations, ranging from small to large scales, and covering both binary and multi-class classification problems.
\item \textbf{Standardized Evaluation Pipeline}: We developed a standardized pipeline for holistically evaluating the quality of the explanations. This pipeline unifies both the implementation of evaluation methods and the format of visual explanations generated by different saliency methods, ensuring seamlessly reproducible experiments, efficient comparisons, and streamlined iteration on existing approaches.
\item \textbf{Extensive Benchmarking and Analysis}: We conducted extensive benchmarking across our dataset collection to assess the quality of the explanations generated by various saliency methods. Our analysis focuses on key evaluation criteria, including alignment and faithfulness, providing insights into the effectiveness of these visual explanations.
\item \textbf{User-Friendly Evaluation Tool}: We offer an easy-to-use tool with an API for querying and accessing our proposed datasets, standardized dataset loading, and performance evaluation, simplifying the evaluation process for researchers. A detailed tutorial for implementation is also provided in the Appendix.
\end{itemize}

\begin{figure*}[ht]
    \centering
    \includegraphics[width=0.9\linewidth]{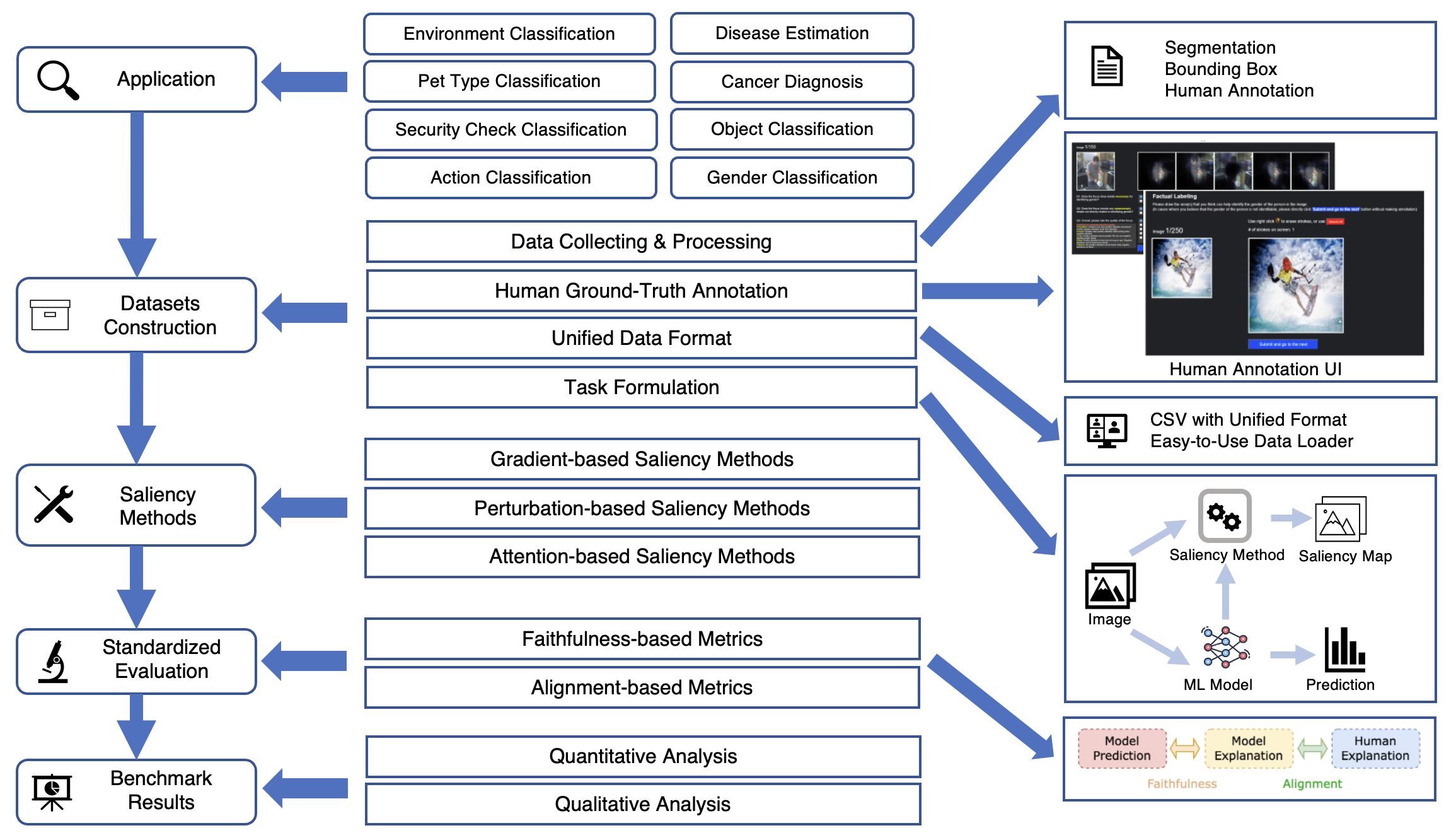}
    \caption{Overview of Saliency-Bench: A Comprehensive Benchmark for Evaluating Visual Explanations.}
    \label{fig:overview}
\end{figure*}

\section{Related Work}
\label{sec:relatedwork}
In this section, we first provide a brief introduction to the primary approaches for saliency methods. Next, we discuss the evaluation metrics commonly used for assessing the performance of saliency methods. Finally, we review existing datasets and benchmarks for XAI and saliency and highlight the gaps that our proposed benchmark aims to address.

\subsection{Saliency Methods for Visual Explanation}
Saliency methods are essential tools for explaining model decisions in image classification tasks by identifying the regions of an image that most influence a model’s prediction. These methods can be broadly classified into two categories: post-hoc explanations and intrinsic explanations.

Post-hoc saliency methods are applied after the model has been trained, providing explanations that highlight important regions in the image based on the model’s output. Gradient-based methods, such as GradCAM~\cite{selvaraju2017grad}, Integrated Gradients~\cite{sundararajan2017axiomatic}, and DeconvNet~\cite{zhou2015learning}, calculate the gradient of the output with respect to the input image, emphasizing regions with the highest gradients. These methods are efficient but can be sensitive to noise in the model and the intricacies of training~\cite{springenberg2014striving}. 

Another class of post-hoc methods includes perturbation-based approaches like RISE~\cite{petsiuk2018rise}, LIME~\cite{mishra2017local}, and MAPLE~\cite{messalas2019model}, which perturb parts of the image and measure how the model’s output changes as a result.

Intrinsic saliency methods, on the other hand, are integrated directly into the model’s architecture, providing explanations from within the model itself. Attention-based methods, such as those used in Vision Transformers~\cite{dosovitskiy2020image} and Swin Transformers~\cite{liu2021swin}, leverage attention mechanisms to reveal which regions of the image the model focuses on during the decision-making process. These methods can be highly informative as they provide direct insight into the model’s internal workings~\cite{fukui2019attention, vaswani2017attention}. 

\subsection{Evaluation Metrics for Saliency Methods}
Various metrics have been proposed to evaluate the alignment and faithfulness of saliency methods, each providing distinct insights into model explanations. \textit{Alignment} metrics gauge how well saliency maps correspond to human understanding of the model's decision-making process. Common alignment metrics include mIoU, Pointing Game~\cite{zhang2018top}, Shared Interest~\cite{boggust2022shared}, MAE Error~\cite{mohseni2021quantitative}, and Human Judgment~\cite{gao2022aligning, selvaraju2017grad, petsiuk2018rise, doshi2017towards}. \textit{Faithfulness} metrics assess how accurately saliency maps reflect the model's true reasoning. Examples of faithfulness metrics include the Insertion Curve (iAUC)~\cite{petsiuk2018rise}, Deletion~\cite{petsiuk2018rise}, AOPC, and Faithfulness F~\cite{tomsett2020sanity}. In addition to alignment and faithfulness, other evaluation methods, such as False-positives~\cite{yang2019benchmarking}, Sensitivity~\cite{guillaumin2012large}, and Stability~\cite{yeh2019fidelity}, have been proposed but are less commonly adopted. Additionally, toolkits such as Xplique~\cite{fel2022xplique}, Captum~\cite{kokhlikyan2020captum}, and Quantus~\cite{hedstrom2023quantus} offer automated implementations and evaluations of XAI methods.

Despite the availability of these metrics, there is still no standardized benchmark, making it difficult to assess and compare saliency methods consistently, especially when ground-truth annotations are lacking.

\subsection{Datasets for XAI and Saliency Benchmarking}
Several datasets have been developed to evaluate saliency methods. CLEVR-XAI~\cite{ARRAS202214} and VQA-HAT~\cite{das2017human} focus on evaluating visual explanations for visual question answering (VQA) tasks, using questions and ground-truth masks. Previous studies have utilized the PASCAL VOC~\cite{chattopadhay2018grad} and ImageNet~\cite{deng2009imagenet} datasets, incorporating multi-layer human attention masks aggregated from multiple annotators to evaluate saliency methods~\cite{mohseni2021quantitative}. Additionally, some works have employed bounding boxes~\cite{boggust2022shared} as a proxy for ground-truth annotations in assessing model explanations. In the medical domain, chest X-ray segmentation datasets, such as CheXpert~\cite{irvin2019chexpert}, provide radiologist-annotated segmentations for evaluating saliency methods on medical imaging tasks~\cite{saporta2022benchmarking}. FunnyBirds~\cite{hesse2023funnybirds} introduces a synthetic vision dataset designed for the automatic and quantitative analysis of XAI methods through image interventions. 

Efforts in other domains have focused on datasets for benchmarking saliency methods in NLP, tabular, and graph data. ERASER~\cite{deyoung2020eraser} provides human-annotated rationales for NLP, with additional textual rationales from forums~\cite{boyd2022human, zaidan2007using, khashabi2018looking}. XAI-Bench~\cite{liu2021synthetic} offers synthetic datasets for feature attribution in tabular data, while OpenXAI~\cite{agarwal2023openxai} provides real-world datasets and metrics. For graph data, SHAPEGGEN~\cite{agarwal2023evaluating} and G-XAI Bench~\cite{agarwal2023evaluating} offer synthetic and real-world datasets for evaluating GNN explainers, and Hruska et al.~\cite{hruska2022ground} introduced a dataset for chemical property prediction on molecular graphs.

Despite existing efforts, unified, large-scale, diverse, and consistently annotated datasets for evaluating saliency methods, particularly in image classification, remain scarce. Our work fills this gap by providing eight diverse image datasets with human-annotated explanations and a standardized evaluation framework.



\section{Task Formulation}
Saliency methods are widely used in XAI to highlight the regions in an image that most influence a model's prediction. These methods generate saliency maps, which visually represent the contribution of each pixel to the model's output. Given an input image \( I \in \mathbb{R}^{c_h \times h \times w} \), where \( c_h \), \( h \), and \( w \) represent the number of channels, height, and width of the image, a black-box classifier can be described by the function \( f: \mathbb{R}^{c_h \times h \times w} \to \mathbb{R}^C \), where \( C \) is the number of classes. A saliency method generates a saliency map \( S = E(I, f) \in \mathbb{R}^{h \times w} \), where the saliency map has the same spatial dimensions as the input image.

Saliency methods provide insights into the model's decision-making by quantifying the relevance of individual pixels in relation to the final output. These methods are evaluated based on their ability to highlight regions that significantly influence the model’s predictions, aligning the model’s rationale with human reasoning.

\section{A Comprehensive Benchmark for Evaluating Visual Explanations}
In this section, we introduce Saliency-Bench, our proposed benchmark for evaluating visual explanations in XAI. We begin with an overview of Saliency-Bench in Section~\ref{sec:overview}, followed by a detailed description of the dataset collection in Section~\ref{sec:datasets}, which includes both an overview of the datasets and the annotation process. Finally, we discuss the proposed standardized evaluation pipeline to assess visual explanation in Section~\ref{sec:evaluation}.

\subsection{Overview of Saliency-Bench}
\label{sec:overview}
To advance research in XAI and saliency methods, we introduce Saliency-Bench, a comprehensive benchmark and dataset collection for evaluating visual explanations, as shown in Figure~\ref{fig:overview}. This framework provides a standardized evaluation paradigm that assesses both the faithfulness and alignment of generated saliency maps. Saliency-Bench comprises a diverse set of datasets across various domains, with sizes ranging from small to large. To ensure consistency and facilitate widespread use, we have unified the data format and developed a modular evaluation pipeline. This standardized pipeline unifies both the implementation of evaluation methods and the format of the visual explanations generated by different saliency methods, ensuring seamlessly reproducible experiments, efficient comparisons, and streamlined iterations on existing approaches. In addition to the benchmarking experiments, we provide an in-depth analysis of the performance of multiple saliency methods, addressing both alignment and faithfulness. Overall, Saliency-Bench offers a scalable, modular, and comprehensive framework for evaluating and improving saliency methods in XAI.

\subsection{Dataset Collection}
\label{sec:datasets}
In constructing our dataset collection, we curated images from eight datasets across diverse domains, including gender classification, object classification, scene recognition, nodule classification, tumor detection, and action classification, each varying in task complexity and annotation methods to enable a comprehensive evaluation of saliency methods. The annotation approaches were tailored to the nature of each task. For some datasets, ground-truth explanations were provided through human annotation, while others utilized foreground extraction techniques to highlight the most relevant regions in the images. We developed specialized user interfaces (UIs) to facilitate the human annotation process, ensuring consistency and high-quality explanations (see the Appendix for a detailed description of the dataset construction and annotation process). Figure~\ref{fig:teaser} presents example images from each dataset. Table~\ref{tab:dataset} offers a summary of the key characteristics of the datasets, including the class types, sizes, and annotation methods used. The following sections provide more specific details on each dataset.

\begin{table*}[ht]
    \centering
    \resizebox{0.8\textwidth}{!}{
\begin{tabular}{l c c c c c c}
\toprule			
Dataset & Class Type & \# of Classes & Size & Annotation Type & Format & Balanced \\
\midrule
\midrule
Gender-XAI      & Binary & 2 & 5,000 & Human annotation & Pixel-wise &\CheckmarkBold \\
Environment-XAI & Binary & 2 & 5,000 & Human annotation & Pixel-wise & \CheckmarkBold \\
Disease-XAI     & Binary & 2 & 5,250 & Human annotation & Pixel-wise & \CheckmarkBold \\
Cancer-XAI & Binary  & 2 & 361 & Human annotation & Pixel-wise & \XSolidBrush \\
Security-XAI & Binary & 2 & 17,654 & Human annotation & Bounding-box & \CheckmarkBold \\
Pet-XAI & Multi-class & 37 & 7,390 & Foreground extraction& Pixel-wise & \XSolidBrush \\
Action-XAI & Multi-class & 127 & 11,511 & Foreground extraction& Pixel-wise & \XSolidBrush \\
Object-XAI & Multi-class & 20 & 4,318 & Foreground extraction& Pixel-wise & \XSolidBrush \\
\bottomrule
\end{tabular}
}
\caption{Summary of Available Datasets: The ``Class Type'' column indicates whether the dataset involves binary or multi-class classification. The ``\# of Classes’’ column shows the total number of categories within each dataset. The ``Size'' column lists the total number of image samples, class labels, and corresponding explanation annotations in each dataset. The ``Annotation Type'' column specifies the source of the explanation annotations, while the ``Format'' column denotes how the annotations are provided. Finally, the ``Balanced'' column indicates whether the dataset maintains class balance for the predictive label.}
\label{tab:dataset}
\end{table*} 

\noindent{\bfseries Gender Classification Dataset (Gender-XAI)}
The gender classification dataset is derived from the Microsoft COCO dataset~\cite{DBLP:journals/corr/LinMBHPRDZ14}. To construct the dataset, we extracted images from the COCO dataset that contained captions with the terms~\textit{man} or~\textit{woman}. Further filtering was performed to remove images that mentioned both genders in the caption, depicted multiple individuals, or featured unrecognizable human figures. Furthermore, a subset of the images was manually annotated by human annotators using human annotation UIs. The dataset comprises a total of 5,000 images with class labels and 3,454 human explanation annotations, evenly distributed between females and males.

\noindent{\bfseries Environment Classification Dataset (Environment-XAI)}
The environment classification dataset used in our study is derived from the Places365 dataset~\cite{zhou2017places} and further annotated manually with human annotation UIs. The task of this dataset involves binary classification for scene recognition, specifically distinguishing between natural and urban scenes. To create the dataset, we selectively sampled images from specific categories. Specifically, the categories used to sample the data are: \textit{Nature}: mountain, pond, waterfall, field wild, forest broadleaf, rainforest; and \textit{Urban}: house, bridge, campus, tower, street, and driveway. In total, the dataset comprises 5,000 images with class labels and 3,052 human annotations.

\noindent{\bfseries Disease Risk Estimation Dataset (Disease-XAI)} We constructed the disease risk estimation dataset from the LIDC-IDRI~\cite{armato2011lung}, which comprises thoracic computed tomography (CT) scans from lung cancer screenings annotated with lesion markers. We converted the 3D nodule images into 2D by selecting the central slice along the z-axis and resizing it to 224×224 pixels. Up to four experienced thoracic radiologists provided annotations in XML format for each scan. The ground truth explanation was established by computing a consensus volume from these annotations, with a nodule considered \textit{positive} if agreed upon by at least 50\% of the radiologists. Conversely, \textit{negative} samples were derived by slicing surrounding areas without nodules. Post-preprocessing, the dataset includes 2,625 \textit{positive} nodule images with human explanation annotations and 2625 \textit{negative} non-nodule images. The primary objective of utilizing this dataset is to determine the presence or absence of nodules.

\noindent{\bfseries Cancer Diagnosis Dataset (Cancer-XAI)} 
We sourced normal pancreas images from the Cancer Imaging Archive~\cite{roth2015deeporgan}. Abnormal scans, featuring pancreatic tumors, were derived from the Medical Segmentation Decathlon dataset (MSD), where initial ground-truth annotations by a medical student were rigorously reviewed and refined by a skilled radiologist. The final cancer diagnosis dataset includes 281 \textit{positive} scans, identified by the presence of tumors, and 80 \textit{negative} scans without tumor indications. In a preprocessing approach akin to that used for the LIDC-IDRI dataset, we converted the 3D scans into 2D slices by randomly selecting along the z-axis, thus setting the stage for a binary classification task to discern between \textit{positive} (tumorous) and \textit{negative} (normal) pancreatic samples.

\noindent{\bfseries Security Check Classification Dataset (Security-XAI)} 
The security check classification task in our study is constructed using the Sixray dataset~\cite{miao2019sixray}. The Sixray dataset, partitioned based on the recognition of prohibited items, comprises an extensive suite of 1,059,231 X-ray images. Each image is annotated at the image level by experienced security inspectors, whose expert annotations we repurpose as human explanation annotations. This approach capitalizes on professional insight, ensuring that our dataset's annotations reflect real-world classification scenarios and provide a reliable basis for the binary classification of prohibited items. Consequently, the dataset has 17,654 images evenly distributed in positive and negative classes and each positive image has a corresponding human explanation annotations.

\noindent{\bfseries Pet Type Classification Dataset (Pet-XAI)} 
The pet type classification dataset used in our study is constructed from The Oxford-IIIT Pet Dataset~\cite{parkhi12a}. This dataset, tailored for pet image analysis, contains over 7,000 images across 37 unique categories, each corresponding to different breeds of dogs or cats. For the purposes of our research, we treat pixel-level foreground extractions, which isolate the pet from the background, as proxies for human explanation annotations. These extractions effectively highlight the subject of interest in alignment with the class label, mirroring the focus areas a human annotator might identify when asked to explain the basis for classifying an image as either a~\textit{dog} or a~\textit{cat}. The resulting dataset contains 2,400 images for~\textit{cat} and 4,990 images for~\textit{dog}, with a total of 7,349 human explanation annotations. 

\noindent{\bfseries Action Classification Dataset (Action-XAI)} The Action Classification Dataset, Action-XAI, is derived from the VQA-based visual and textual explanations dataset, Activity Explanation (ACT-X)~\cite{park2018multimodal,zhang2024megl}. We extracted samples from the ACT-X dataset that could be converted from a VQA task into a classification task. To determine whether a question-answer pair could be restructured into a classification problem, we employed LLMs to verify the suitability of each pair for conversion. The resulting Action-XAI dataset is designed for object classification, where each image sample includes visual explanations that justify the assigned class label. The dataset consists of 127 distinct action class labels, providing a diverse set of actions for classification tasks. In total, the dataset contains 11,511 image samples.

\noindent{\bfseries Object Classification Dataset (Object-XAI)} The Object Classification Dataset utilized in our study is constructed from the PASCAL VOC 2012 Dataset~\cite{Everingham10}. The dataset comprises roughly 11,540 images and covers 20 diverse object categories. These include~\textit{aeroplanes, bicycles, birds, boats, bottles, buses, cars, cats, chairs, cows, dining tables, dogs, horses, motorbikes, people, potted plants, sheep, sofas, trains, and TV monitors}. Each image in the VOC 2012 Dataset is manually annotated with pixel-level region masks and corresponding class labels for identified objects, providing a robust resource for object classification research. In our study, explanation annotations were generated by extracting the pixel-level foreground corresponding to the image label class, leveraging these precise regions as effective proxies for human explanations by directly highlighting the areas most relevant to the object's classification. The finalized dataset contains 4,318 images, each with a class label and corresponding human explanation annotations.

\subsection{Standardized Evaluation Pipeline}
\label{sec:evaluation}
Saliency-Bench adopts a standardized evaluation pipeline that incorporates a broad range of quantitative metrics for assessing explanation quality. It integrates both alignment-based and faithfulness-based metrics. Our pipeline standardizes the evaluation process by unifying both the implementation of these evaluation methods and the format of the visual explanations generated by different saliency methods. This ensures seamlessly reproducible experiments, efficient comparisons, and streamlined iterations on existing approaches. Below, we provide a detailed overview of these metric categories and their role within our comprehensive evaluation framework.

\subsubsection{Alignment-based metrics}
Alignment-based metrics evaluate how well the generated visual explanation aligns with ground-truth explanation annotations. Common alignment metrics include mean Intersection over Union (mIoU) and the Pointing Game (PG). 

\noindent{\bfseries mIoU} To calculate mIoU, the generated saliency map \( E \) is first converted into a binary map \( B \in \{0, 1\}^{m \times n} \) for each sample, based on a threshold \( \theta \). Each pixel in \( B \) is assigned a value of 1 if the corresponding value in \( E \) exceeds \( \theta \), 0 otherwise. The mIoU is then calculated by comparing this binary map with the binary human explanation annotations \( A \), with the formula:
\[
\text{mIoU}(E, A) = \frac{1}{N} \sum_{i=1}^N \frac{|B_i \cap A_i|}{|B_i \cup A_i|},
\]
where \( N \) is the number of samples and \( B_i \) and \( A_i \) are the binary maps for the \( i \)-th sample.

\noindent{\bfseries PG} The Pointing Game~\cite{zhang2018top} evaluates whether the peak of the saliency map \( E \) for each sample falls within the human-annotated explanation region \( A \). It is defined as:
\[
\text{Pointing Game} = \frac{\sum_{i=1}^N 1[\operatorname{MaxLoc}(E_i) \in A_i]}{N},
\]
where \( N \) is the number of samples, and \( \operatorname{MaxLoc}(E_i) \) represents the location of the highest activation in the saliency map \( E \) for the \( i \)-th sample.

\begin{figure}[ht]
    \centering
    \includegraphics[width=0.9\linewidth]{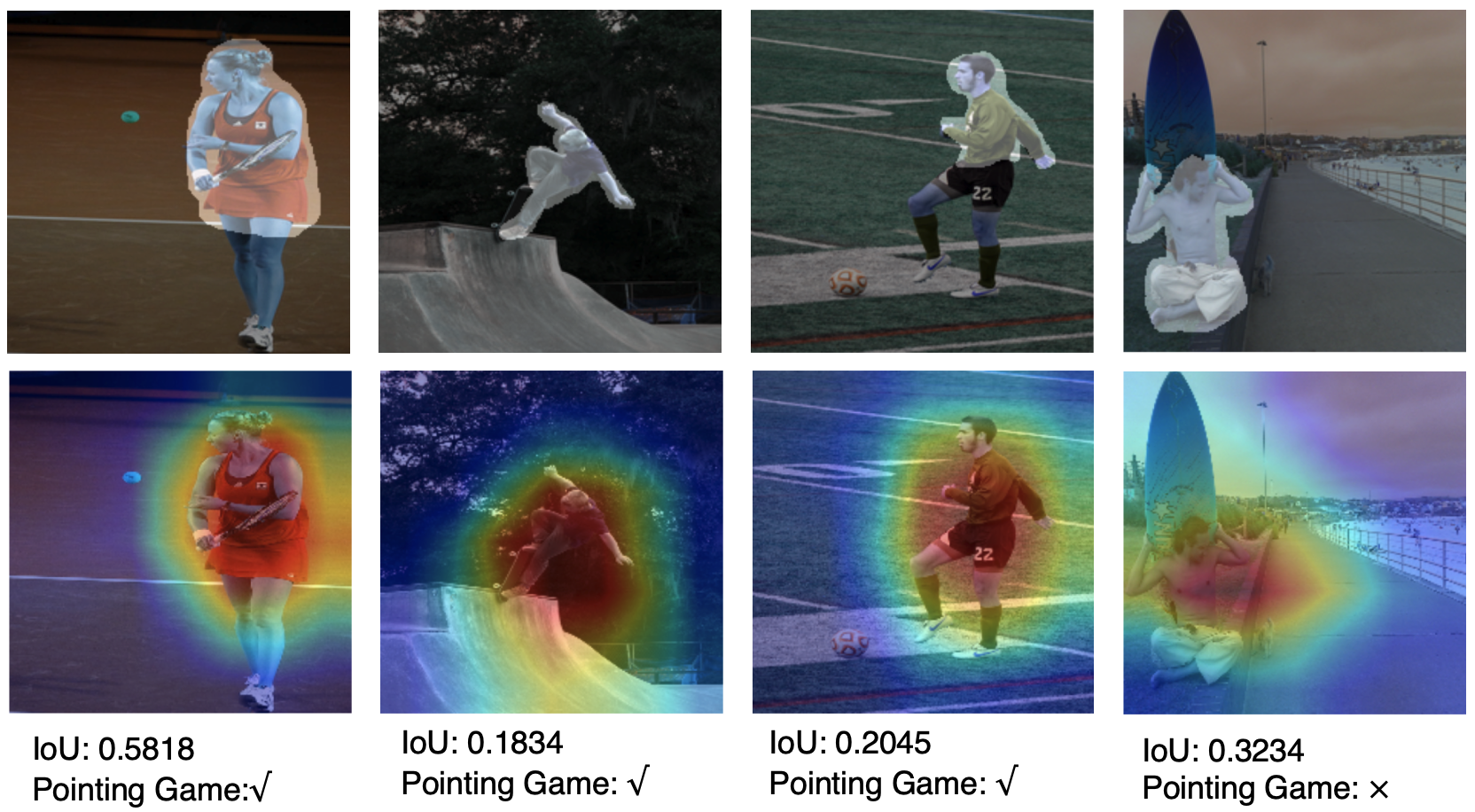}
    \caption{Examples of mIoU and Pointing Game comparing saliency maps generated by Grad-CAM with ground-truth annotations on the Action-XAI dataset.}
    \label{fig:example_alignment}
\end{figure}

An example of mIoU and PG evaluations is shown in Figure~\ref{fig:example_alignment}.

\subsubsection{Faithfulness-based metrics}
Faithfulness-based metrics evaluate the causal influence of highlighted features of the visual explanation on the model's output. These metrics test whether manipulating the input regions identified as important by the explanation leads to predictable changes in the model's predictions. The objective is to determine whether the highlighted features are causally linked to the model's output. In this paper, we implement the iAUC metric to assess faithfulness.

\noindent{\bfseries Insertion (iAUC)}: The Insertion metric~\cite{petsiuk2018rise} evaluates how much the model's prediction confidence increases as the most important features, as identified by the explanation, are progressively inserted into a blank image. Starting with an image that contains no features, the most important parts of the image are gradually added. If the explanation is accurate, inserting the highlighted features should significantly boost the model's prediction confidence. Formally, the iAUC is calculated as:
\[
\text{iAUC} = \int_0^1 P(y | I_{\text{insert}}(r)) \, dr,
\]
where \( P(y | I_{\text{insert}}(r)) \) represents the probability of predicting class \( y \) given the input image \( I \), with a fraction \( r \) of the most important pixels inserted.

\section{Experiments}
In this section, we present our benchmarking of state-of-the-art saliency methods using the proposed datasets. We begin by describing the detailed experimental settings in Section~\ref{sec:exp_setting}, followed by a comprehensive evaluation and analysis of the results in Section~\ref{sec:exp_results}.

\subsection{Experimental Settings}
\label{sec:exp_setting}
\noindent{\bfseries Baselines} We benchmark six common saliency methods: GradCAM~\cite{selvaraju2017grad}, GradCAM++~\cite{chattopadhay2018grad}, Integrated Gradients~\cite{sundararajan2017axiomatic}, InputXGradient (IxG)~\cite{shrikumar2016not}, Occlusion~\cite{zeiler2014visualizing}, and RISE~\cite{petsiuk2018rise} in our eight proposed datasets. The experiments were conducted using two different CNN architectures: ResNet-18~\cite{he2016deep} and VGG-19~\cite{simonyan2014very}. We also benchmark the attention mechanism as a saliency explanation method of ViT with ViT-base-patch16-224 architecture (ViT-B/16)~\cite{dosovitskiy2020image}.

\noindent{\bfseries Evaluation metrics} We evaluate the performance of the saliency methods using three metrics: mIoU and PG for alignment-based evaluation and iAUC for faithfulness-based evaluation. For all metrics, the overall results are obtained as the average through all the samples in the test set.

\noindent{\bfseries Implementation Details} In each experiment, we allocated the data as follows: 70\% for training, 15\% for validation, and the remaining 15\% for testing. Both pretrained ResNet-18 and VGG-19 models were trained for 30 epochs using the Adam optimizer with a learning rate of 0.001, and the best checkpoint was selected based on performance on the validation set. The implementation of the saliency methods followed the original settings described in the respective papers. Our experiments are implemented based on PyTorch and performed on an NVIDIA A10G GPU.

\begin{table*}[ht]
\centering
\resizebox{\linewidth}{!}{%
\begin{tabular}{c|c|ccc|ccc|ccc|ccc|ccc|ccc|ccc|ccc|ccc}
\toprule
\multirow{2}{*}{Base Model} & \multirow{2}{*}{Dataset} & \multicolumn{3}{c|}{GradCAM} & \multicolumn{3}{c|}{GradCAM++} & \multicolumn{3}{c|}{Integrated Gradients} & \multicolumn{3}{c|}{IxG} & \multicolumn{3}{c|}{Occlusion} & \multicolumn{3}{c|}{RISE} \\
\cmidrule(r){3-5} \cmidrule(r){6-8} \cmidrule(r){9-11} \cmidrule(r){12-14} \cmidrule(r){15-17} \cmidrule(l){18-20}
                            &                &  mIoU  &   PG   &  iAUC  &  mIoU  &   PG   &  iAUC  &  mIoU  &   PG   &  iAUC  &  mIoU  &  PG    &  iAUC  &  mIoU  &  PG    & iAUC   &  mIoU  &   PG   &  iAUC  \\
\hline
\hline
\multirow{9}{*}{ResNet-18}  & Gender-XAI     & 0.1304 & 0.3077 & 0.6880 & 0.1417 & 0.2906 & 0.6625 & 0.3441 & 0.2138 & 0.3628 & 0.3403 & 0.2911 & 0.4011 & 0.1301 & 0.2618 & 0.6787 & 0.3416 & 0.3307 & 0.8454 \\
                            & Environment-XAI& 0.3656 & 0.4777 & 0.6911 & 0.3632 & 0.4224 & 0.7009 & 0.5356 & 0.2810 & 0.5114 & 0.4566 & 0.3010 & 0.5722 & 0.3812 & 0.4596 & 0.6778 & 0.5428 & 0.3624 & 0.9230 \\
                            & Disease-XAI    & 0.0967 & 0.2354 & 0.6017 & 0.1028 & 0.2501 & 0.6154 & 0.1254 & 0.1714 & 0.5297 & 0.1389 & 0.2802 & 0.4365 & 0.1497 & 0.2901 & 0.6401 & 0.1260 & 0.2705 & 0.6334 \\
                            & Cancer-XAI     & 0.0620 & 0.2105 & 0.5523 & 0.0654 & 0.2207 & 0.5645 & 0.0913 & 0.1385 & 0.4732 & 0.1049 & 0.2325 & 0.4806 & 0.1107 & 0.2394 & 0.5892 & 0.1153 & 0.2451 & 0.5901 \\
                            & Security-XAI   & 0.1243 & 0.6984 & 0.6136 & 0.1292 & 0.6120 & 0.6258 & 0.1420 & 0.6225 & 0.5322 & 0.1552 & 0.6305 & 0.4350 & 0.1667 & 0.6420 & 0.6424 & 0.1503 & 0.6301 & 0.6398 \\
                            & Pet-XAI        & 0.1564 & 0.3215 & 0.7596 & 0.1631 & 0.3352 & 0.7694 & 0.1752 & 0.2465 & 0.5799 & 0.1790 & 0.3583 & 0.5451 & 0.1820 & 0.3675 & 0.7894 & 0.1855 & 0.3738 & 0.7924 \\
                            & Action-XAI     & 0.4009 & 0.3982 & 0.6341 & 0.4136 & 0.3989 & 0.6462 & 0.4214 & 0.3068 & 0.6578 & 0.4300 & 0.4143 & 0.6702 & 0.4351 & 0.4202 & 0.6750 & 0.4406 & 0.4295 & 0.6853 \\
                            & Object-XAI     & 0.4960 & 0.2156 & 0.6620 & 0.5084 & 0.2224 & 0.6722 & 0.5153 & 0.3303 & 0.7210 & 0.5265 & 0.2406 & 0.6894 & 0.5350 & 0.2502 & 0.6985 & 0.5430 & 0.2551 & 0.7070 \\
\hline
    \multirow{9}{*}{VGG-19} & Gender-XAI     & 0.1194 & 0.2823 & 0.6102 & 0.1311 & 0.2859 & 0.5823 & 0.2920 & 0.1986 & 0.3776 & 0.1814 & 0.2607 & 0.3890 & 0.1235 & 0.2450 & 0.5323 & 0.2961 & 0.3031 & 0.8089 \\
                            & Environment-XAI& 0.3295 & 0.4174 & 0.6305 & 0.3160 & 0.3871 & 0.6412 & 0.4683 & 0.2629 & 0.4810 & 0.4025 & 0.2897 & 0.5604 & 0.3322 & 0.4219 & 0.6215 & 0.4792 & 0.3497 & 0.8852 \\
                            & Disease-XAI    & 0.0789 & 0.2150 & 0.5801 & 0.0870 & 0.2255 & 0.5452 & 0.1105 & 0.1493 & 0.3570 & 0.1350 & 0.2595 & 0.3678 & 0.1223 & 0.2656 & 0.6279 & 0.1289 & 0.2662 & 0.6120 \\
                            & Cancer-XAI     & 0.0513 & 0.1855 & 0.4701 & 0.0560 & 0.1937 & 0.4851 & 0.0803 & 0.2079 & 0.3340 & 0.0901 & 0.2253 & 0.4524 & 0.1280 & 0.2242 & 0.5600 & 0.1023 & 0.2721 & 0.5807 \\
                            & Security-XAI   & 0.1139 & 0.6765 & 0.5720 & 0.1190 & 0.6876 & 0.5850 & 0.1313 & 0.6470 & 0.4955 & 0.1410 & 0.5058 & 0.4632 & 0.1505 & 0.6153 & 0.6120 & 0.1448 & 0.6184 & 0.6284 \\
                            & Pet-XAI        & 0.1509 & 0.2767 & 0.7302 & 0.1486 & 0.2853 & 0.6920 & 0.1675 & 0.2642 & 0.5010 & 0.1615 & 0.3050 & 0.5298 & 0.1670 & 0.3193 & 0.6450 & 0.1720 & 0.3242 & 0.7110 \\
                            & Action-XAI     & 0.3548 & 0.3642 & 0.5947 & 0.3561 & 0.3755 & 0.5896 & 0.3674 & 0.2742 & 0.5924 & 0.3739 & 0.3802 & 0.6604 & 0.3793 & 0.3850 & 0.7189 & 0.3855 & 0.3951 & 0.6591 \\
                            & Object-XAI     & 0.4319 & 0.2030 & 0.5892 & 0.4812 & 0.2113 & 0.5985 & 0.4526 & 0.4110 & 0.7081 & 0.4631 & 0.2302 & 0.6436 & 0.4730 & 0.2294 & 0.7258 & 0.4831 & 0.2192 & 0.6553 \\
\bottomrule
\end{tabular}%
}
\caption{Evaluation of saliency methods for ResNet-18 and VGG-19.}
\label{tab:method_benchmark}
\end{table*}

\subsection{Results and Analysis}
\label{sec:exp_results}
In this section, we present a detailed analysis of the results obtained from our benchmarking experiments, as summarized in Table~\ref{tab:method_benchmark}.

\subsubsection{Results and Analysis for ResNet-18}
From the results for ResNet-18, we observe a generally strong performance across all datasets. Notably, for the Gender-XAI dataset, the best performance in terms of mIoU is achieved by the Integrated Gradients (0.3441), followed closely by RISE (0.3416). The Occlusion method shows relatively lower performance with a mIoU of 0.1301, indicating that its effectiveness in identifying the regions relevant for model predictions is lower than the other methods. For the Environment-XAI dataset, RISE also outperforms the other methods, yielding a mIoU of 0.5428. This is consistent with the findings from GradCAM++, which ranks second in performance. In contrast, Integrated Gradients shows slightly weaker performance in this case, especially in terms of PG (0.2810), which suggests that its ability to highlight important regions of an image could be improved. The Action-XAI and Object-XAI datasets show that RISE consistently provides robust performance in mIoU (0.4406 and 0.5430, respectively), followed by GradCAM and GradCAM++. Disease-XAI and Cancer-XAI datasets display a notable performance gap, where GradCAM tends to perform better in identifying salient regions compared to the other methods, suggesting that GradCAM's gradient-based approach works better for these types of medical images.

\subsubsection{Results and Analysis for VGG-19}
The results for VGG-19, while showing similar trends, reveal that VGG-19 generally performs slightly worse than ResNet-18 across most datasets, which is consistent with the performance gap typically observed between these two architectures. The Gender-XAI dataset shows that RISE remains the top performer in terms of mIoU (0.2961), though GradCAM++ (0.2859) performs comparably, indicating that the additional information from the gradients in GradCAM++ may be more useful in this case. For Environment-XAI, RISE again excels (mIoU of 0.4792), and GradCAM++ performs slightly worse than ResNet-18 at 0.3160. However, RISE's consistently strong performance across datasets reinforces its utility in identifying salient regions for model predictions. The Disease-XAI and Cancer-XAI datasets present challenges for the saliency methods, with lower mIoU values across the board, particularly for Integrated Gradients (0.1105 for Disease-XAI and 0.0803 for Cancer-XAI), suggesting that the method may struggle with certain types of medical images. In contrast, GradCAM and RISE show more reliable results for these datasets.

\subsubsection{Comparative Analysis}
When comparing ResNet-18 and VGG-19, it is evident that ResNet-18 tends to outperform VGG-19 across most datasets, as indicated by higher mIoU, PG, and iAUC values. This is particularly noticeable for the Action-XAI, Object-XAI, and Environment-XAI datasets, where ResNet-18 consistently leads in terms of both alignment and faithfulness metrics. This difference in performance can likely be attributed to the architecture of the models, with ResNet-18 being a deeper network with skip connections that may help it capture more complex features relevant to saliency mapping. Interestingly, the GradCAM++ method shows stronger performance in VGG-19 compared to ResNet-18 on the Gender-XAI dataset, possibly due to the deeper layers in VGG-19 focusing more effectively on the critical regions of images in gender classification tasks. However, RISE and GradCAM remain the most reliable across the board, showing consistent performance on multiple datasets, regardless of the underlying architecture.


\begin{table*}[ht]
\centering
\resizebox{0.8\linewidth}{!}{
\begin{tabular}{l|l|cccccc}
\toprule
Base Model & Correlation & GradCAM & GradCAM++ & Integrated Gradients & IxG & Occlusion & RISE \\
\hline
\hline
\multirow{3}{*}{ResNet-18} & mIoU \& PG      & 0.3718 & 0.3285 & 0.8462 & 0.0924 & 0.3435 & 0.2994 \\
                           & mIoU \& iAUC    & 0.2822 & 0.3328 & 0.3630 & 0.7246 & 0.2592 & 0.5932 \\
                           & PG \& iAUC      & 0.4623 & 0.4919 & 0.7024 & 0.2366 & 0.3402 & 0.4566 \\
\hline
\multirow{3}{*}{VGG-19}    & mIoU \& PG      & 0.4325 & 0.3290 & 0.6663 & 0.2119 & 0.3200 & 0.1464 \\
                           & mIoU \& iAUC    & 0.2416 & 0.3692 & 0.6722 & 0.8287 & 0.8177 & 0.5840 \\
                           & PG \& iAUC      & 0.4409 & 0.5489 & 0.9312 & 0.4323 & 0.3025 & 0.3743 \\
\bottomrule
\end{tabular}
}
\caption{Pairwise correlation coefficients between evaluation metrics for different saliency methods across two base models, ResNet-18 and VGG-19.}
\label{tab:cc}
\end{table*}

\begin{table}[ht]
\centering
\resizebox{0.8\linewidth}{!}{%
\begin{tabular}{l|cccc}
\toprule
Dataset & mIoU & PG & iAUC \\
\hline
\hline
Gender-XAI      & 0.3698 & 0.3802 & 0.6713 \\
Environment-XAI & 0.5263 & 0.3857 & 0.7632 \\
Disease-XAI     & 0.1568 & 0.3052 & 0.6874 \\
Cancer-XAI      & 0.1330 & 0.2851 & 0.6605 \\
Security-XAI    & 0.2044 & 0.7657 & 0.6826 \\
Pet-XAI         & 0.3056 & 0.2729 & 0.8012 \\
Action-XAI      & 0.3351 & 0.3424 & 0.6798 \\
Object-XAI      & 0.5116 & 0.3158 & 0.7197 \\
\bottomrule
\end{tabular}
}
\caption{Benchmarking results for the ViT-B/16 attention mechanism: mIoU, PG, and iAUC scores on eight datasets.}
\label{tab:vit_attention_benchmark}
\end{table}

\subsubsection{Inter-method reliability analysis}
Inter-method reliability analysis examines the consistency of evaluation metrics across different saliency methods, providing insights into whether the metrics yield stable and comparable assessments of explanation quality. We performed this analysis to assess the correlation between mIoU, PG, and iAUC metrics for two base models, ResNet-18 and VGG-19. The results in Table~\ref{tab:cc} show generally positive correlations across all metrics, with stronger associations observed for ResNet-18. Notably, mIoU and PG have the highest correlation for Integrated Gradients (0.8462), indicating good alignment between the saliency maps and ground-truth annotations. Similarly, mIoU and iAUC are strongly correlated, especially for Integrated Gradients (0.7024). For VGG-19, the correlations are slightly weaker, with GradCAM++ showing the highest PG and iAUC correlation (0.5489). These results suggest that different saliency methods exhibit varying levels of alignment and faithfulness, with some methods (e.g., GradCAM++) being more consistent across metrics than others.


\subsubsection{Benchmarking on Vision Transformer}
In this section, we present the benchmarking results of saliency methods using the ViT. Specifically, we employed the ViT-base-patch16-224 architecture~\cite{dosovitskiy2020image} (ViT-B/16) and focused on evaluating the attention mechanism as a saliency explanation method. The results, presented in Table~\ref{tab:vit_attention_benchmark}, include evaluations of mIoU, PG, and iAUC across eight datasets.

\begin{figure}[ht]
    \centering
    \includegraphics[width=0.8\linewidth]{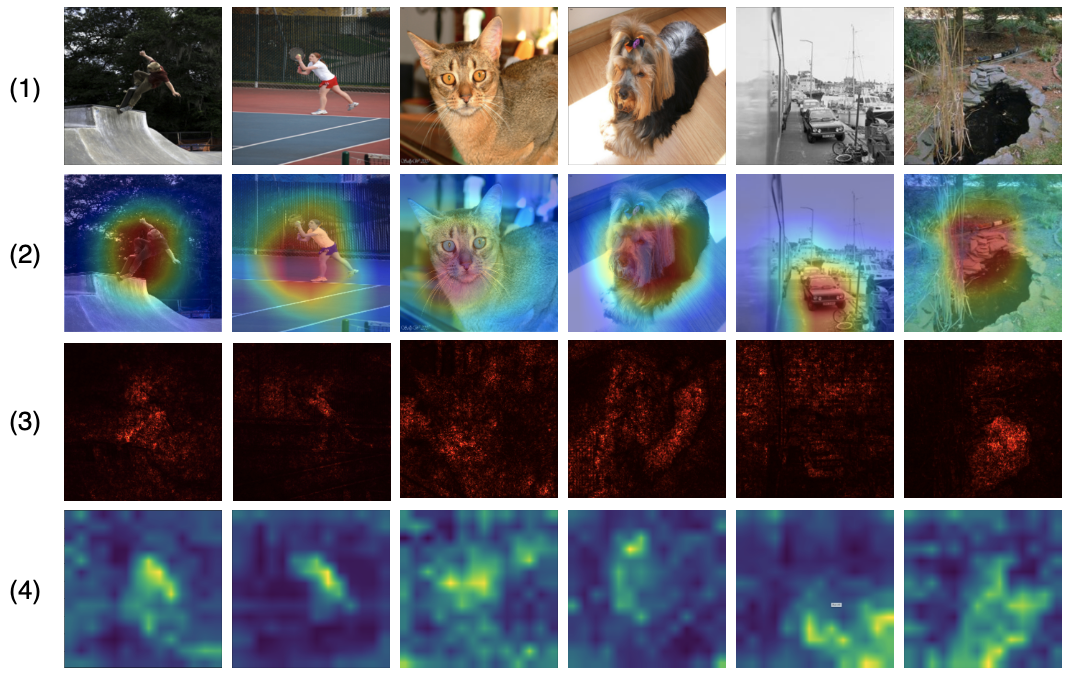}
    \caption{Qualitative results of visual explanation methods: (1) Original image; (2) Saliency map generated by GradCAM; (4) Saliency map generated by InputXGradient; (3) Generated by attention mechanisms of ViT-B/16.}
    \label{fig:example_visual_comparison}
\end{figure}

As illustrated in Figure~\ref{fig:example_visual_comparison}, generates attention maps based on its transformer architecture, which we directly analyze in this benchmarking. The attention maps produced by ViT are capable of capturing more detailed features of regions of interest compared to traditional saliency methods based on CNNs such as GradCAM and IxG. These maps show a more refined focus on relevant areas, leading to improved performance in the evaluation metrics. This improved performance can be attributed to ViT’s self-attention mechanism, which allows the model to attend to long-range dependencies and contextual information within the image, providing a more holistic view of the important regions compared to methods that rely solely on local pixel-level information.

\subsection{Observation and Discussion}
(1) The performance of saliency methods varies across datasets, influenced by factors such as task complexity, label distribution, and image characteristics. Simpler tasks generally yield better results, while more complex datasets tend to present challenges for some methods. (2) ViT’s attention mechanism emerges as a promising alternative to CNN-based saliency methods, especially for capturing global context and long-range dependencies. However, its robustness and efficiency relative to traditional CNN-based approaches require further investigation. (3) The inter-method reliability analysis highlights varying correlations between evaluation metrics, indicating that saliency methods show different levels of consistency across tasks and models. This emphasizes the importance of careful method selection depending on the specific evaluation goals. (4) The analysis suggests that more complex datasets often lead to reduced faithfulness in saliency methods. This indicates that incorporating additional context or information, such as object-level segmentation, could enhance alignment and faithfulness, particularly in challenging tasks. (5) Overall, while the evaluated methods provide valuable insights, each has its strengths and limitations. Improvements are needed, particularly for more intricate tasks requiring fine-grained explanations.

\section{Conclusion and Limitations}
We introduce Saliency-Bench, a comprehensive benchmark suite for evaluating visual explanations generated by saliency methods in image classification. Our benchmark includes eight diverse datasets spanning gender classification, environment classification, action classification, object classification, cancer diagnosis, disease estimation, pet type classification, and security check classification, each with ground-truth explanations. We conduct extensive benchmarking experiments using six widely adopted saliency methods evaluating them with multiple performance metrics, including mIoU, Pointing Game, and iAUC. These methods are tested across different image classifier architectures, including ResNet-18 and VGG-19, providing a comprehensive analysis of their performance. Additionally, we explore ViT-B/16 as a saliency method and perform an inter-method reliability analysis. To facilitate future research, we provide an easy-to-use API for dataset loading, saliency map generation, and evaluation, streamlining the benchmarking process. By standardizing the evaluation of visual explanations, Saliency-Bench aims to drive progress in XAI.

Despite its contributions, our benchmark has limitations. Human-annotated explanations in datasets like Gender-XAI and Scene-XAI may still introduce biases despite independent assessments. The pancreatic tumor detection dataset combines samples from different sources, which could cause unintended dataset biases. Additionally, the inclusion of gender classification may raise ethical concerns related to reinforcing gender stereotypes. These challenges underscore the need for continued refinement in dataset construction and evaluation methodologies.


\bibliographystyle{ACM-Reference-Format}
\bibliography{main}

\newpage
\appendix

In the appendix, we provide an introduction and tutorial on how to utilize our developed user-friendly API of Saliency-Bench. Our API is powered by the Python library named ~\textit{xaibenchmark} and provides users with the ability to leverage our published datasets and evaluate results. This tutorial guides users through the process of conveniently loading the datasets and conducting evaluations on visual explanations generated by saliency methods using both alignment and faithfulness-based metrics. 

\section{Benchmark Usage}
\subsection{Installation}
\begin{figure}[h]
\centering
\includegraphics[width=\linewidth]{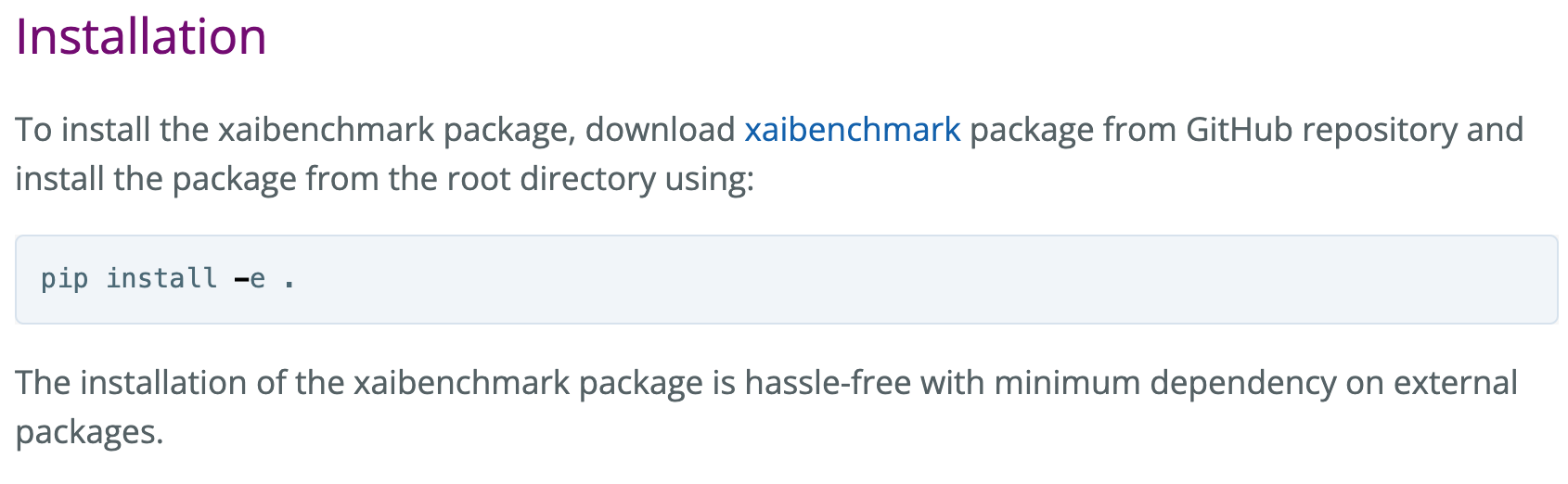}
\caption{\textit{xaibenchmark} Python package installation.}
\label{fig:installation}
\end{figure}

The first step for utilizing our proposed benchmark is to download and install our developed~\textit{xaibenchmark} python package for \textit{saliency-bench}, as shown in Figure~\ref{fig:installation}. 

\subsection{Download and load a dataset}
\begin{figure}[h]
\centering
\includegraphics[width=\linewidth]{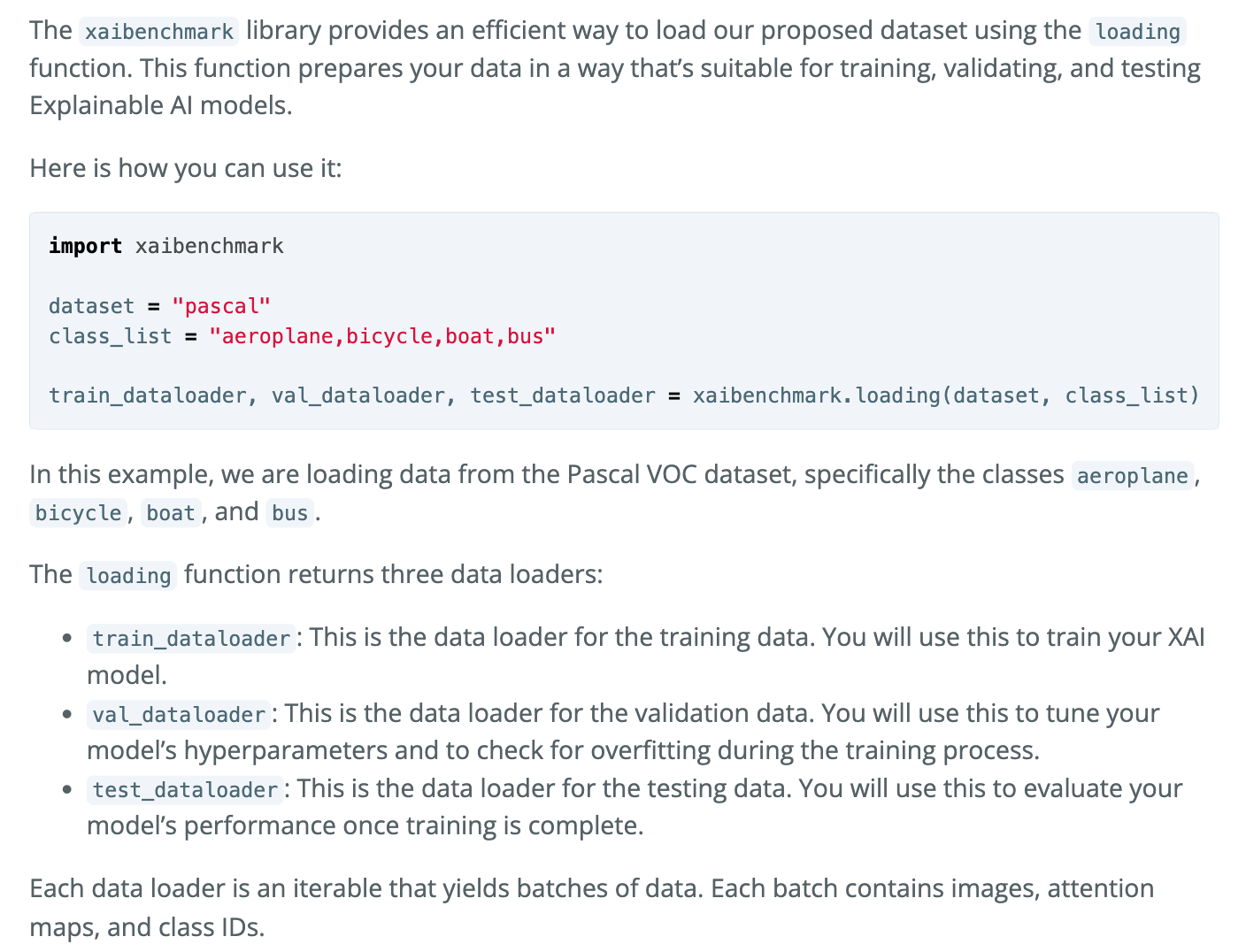}
\caption{Prepare data splits with DataLoaders.}
\label{fig:dataloader}
\end{figure}

We first need to download the dataset from Google Drive and unzip it. The dataset can then be loaded from Python with DataLoaders, as shown in Figure~\ref{fig:dataloader}.

\subsection{Iterate through the DataLoader}
\begin{figure}[h]
\centering
\includegraphics[width=\linewidth]{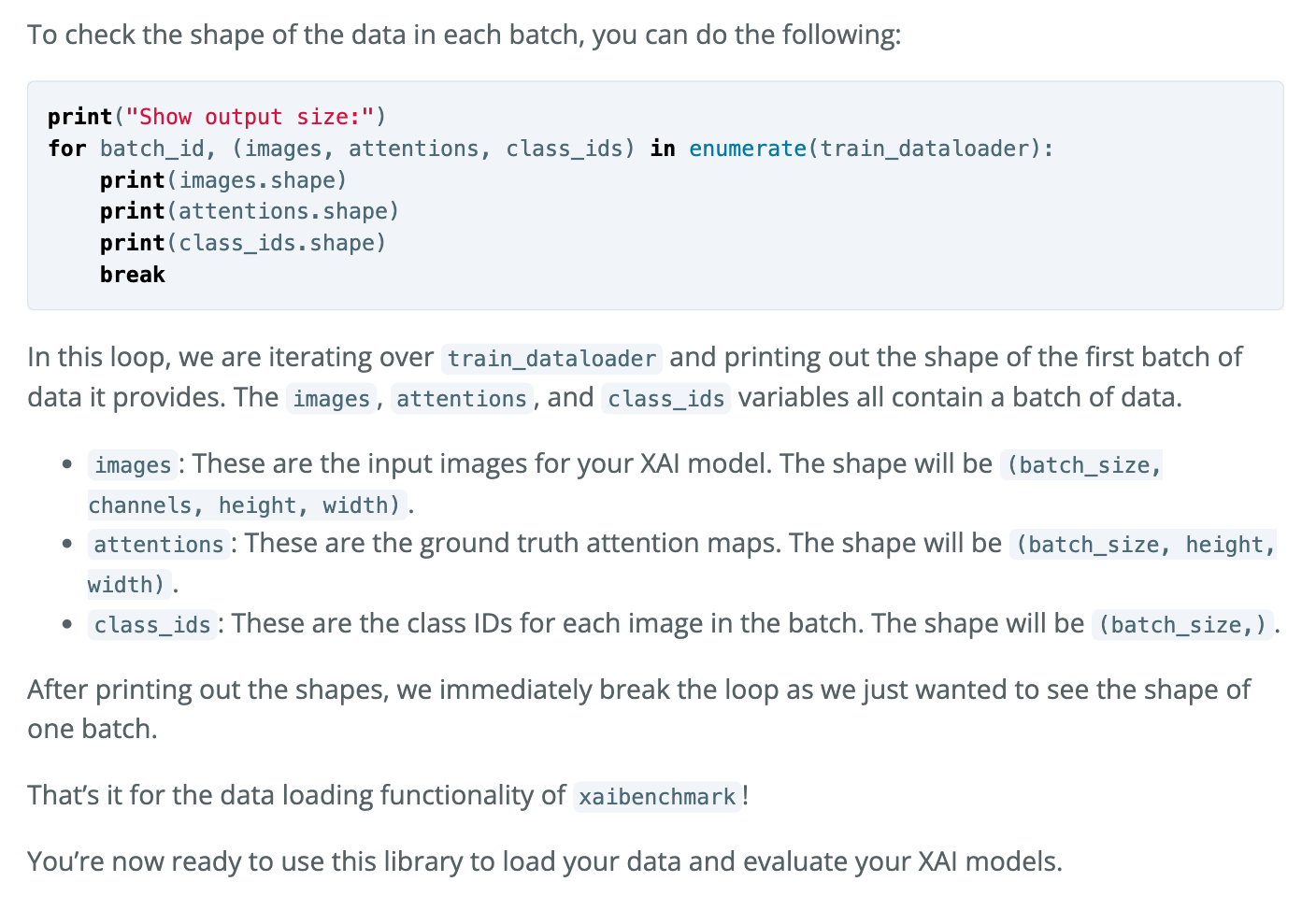}
\caption{Iterate data through the DataLoader.}
\label{fig:load_data}
\end{figure}

We can iterate through the loaded dataset using dataloaders. By iterating over these dataloaders, we can efficiently pass images, human explanation annotations, and class IDs (encoded from class names) from the dataset in ``minibatches,'' as shown in Figure~\ref{fig:load_data}.

\subsection{Evaluation Usage}
\begin{figure}[h]
\centering
\includegraphics[width=\linewidth]{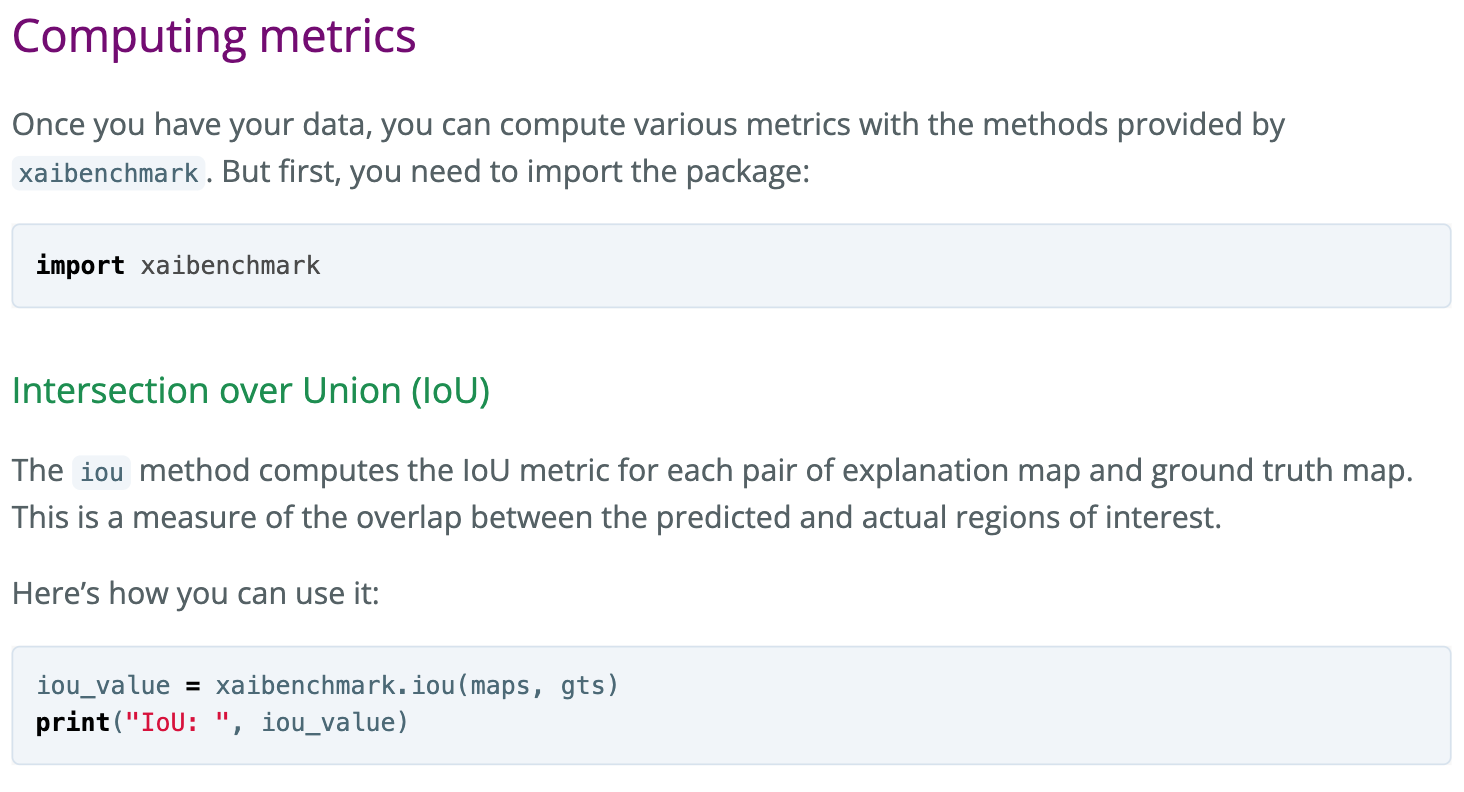}
\caption{Compute visual explanation evaluation metrics.}
\label{fig:evaluation}
\end{figure}

Once we want to evaluate a saliency method, we can use it to generate visual explanations on our published datasets. Subsequently, various evaluation metrics, using IoU as an example shown in Figure~\ref{fig:evaluation}, can be computed against ground-truth human explanation annotations with our benchmark.

\section{Human Annotation User Interface (UI)}

We developed two user interfaces (UIs): Human Explanation Annotation UI and Explanation Qualities Assessment UI. These interfaces facilitate the annotation of human explanation and the assessment of explanation qualities~\cite{gao2022res, gao2022aligning}.

\begin{figure}[h]
\centering
\includegraphics[width=\linewidth]{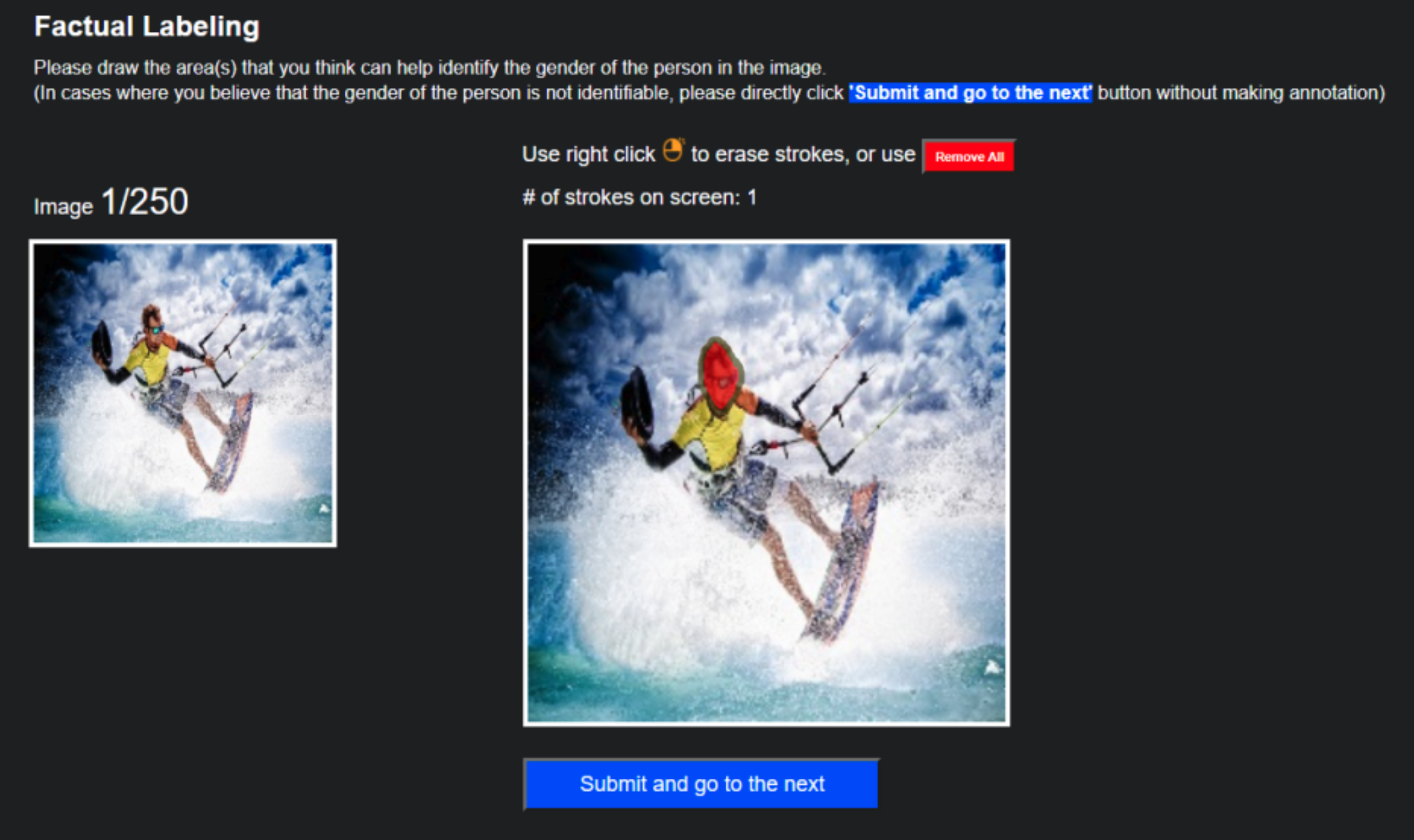}
\caption{UI for Human Explanation Annotation.}
\label{fig:ui_labeling}
\end{figure}

\noindent{\bfseries Human Explanation Annotation UI} Figure~\ref{fig:ui_labeling} illustrates the interface used to collect human explanation annotations by capturing regions that humans intuitively associate with classification decisions. Users can draw directly on the image to create a binary mask of the regions, which serves as a candidate for the human explanation annotation. 

\begin{figure}[h]
\centering
\includegraphics[width=\linewidth]{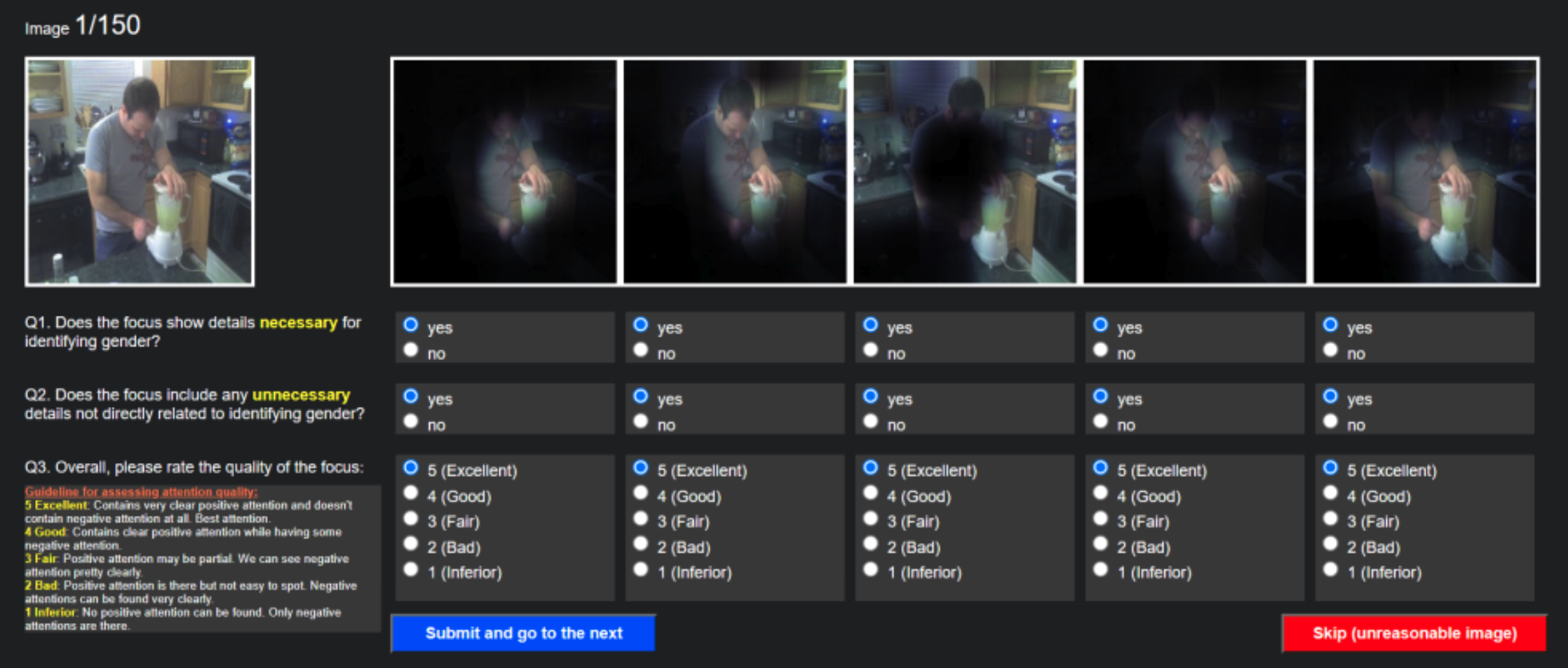}
\caption{UI for Explanation Qualities Assessment.}
\label{fig:ui_evaluation}
\end{figure}

\noindent{\bfseries Annotation Qualities Assessment UI} Figure~\ref{fig:ui_evaluation} shows the interface for assessing the quality of model-generated explanations. Five model-generated explanations are presented in random order, and users are prompted to answer three questions about each to evaluate their quality. This UI is instrumental in evaluating the explainability of visual explanation methods based on human rationale.

\section{Dataset Construction and Annotation Process}
The construction and annotation of the datasets were tailored to the nature of each task, ensuring robust and high-quality ground-truth annotations for model evaluation.

For the Gender-XAI and Environment-XAI datasets, we involved five human annotators who used the annotation UI, as shown in Figure~\ref{fig:ui_labeling}, to mark regions they considered relevant to the classification task. Annotators were instructed to highlight key areas that they believed contributed to the classification decision. To ensure the annotations were of high quality and free from individual biases, the marked regions were reviewed by multiple evaluators with the quality assessment UI, as shown in Figure~\ref{fig:ui_evaluation}.

For the Disease-XAI and Cancer-XAI datasets, we processed the images and aggregated the annotations from four experienced radiologists, who independently annotated the images in the original datasets. To ensure accuracy and minimize subjectivity, the final annotation was determined through a consensus process.

The annotation process was more straightforward for other datasets with binary class labels, such as Security-XAI. We obtained segmentation masks for the positive class (e.g., prohibited items), but no annotations were made for the negative class, as the evaluation focused solely on the positive class.

For the multi-label datasets, we employed a filtering strategy to ensure focused annotations. In Object-XAI, we filtered images to include only those containing a single object class, excluding images with multiple object classes, so that each image focused on a single object for classification. After filtering, we extracted segmentation masks corresponding to the object of interest, which were then used as ground-truth annotations for evaluation. For Pet-XAI, we retained images containing only one dog or one cat and labeled both the pet class and the specific species. For the Action-XAI dataset, the filtering process was more complex. We first used LLMs to determine whether a VQA sample could be converted into a classification task. This conversion depended on whether the question-answer pair could be interpreted as a clear classification problem, such as identifying a specific action. We prompted GPT-4o mini via the OpenAI API to assess the suitability of each pair for conversion. After filtering, we combined the results with ground-truth annotations from the ACT-X dataset to construct Action-XAI. Here is the prompt used to query GPT-4o mini:
\begin{quote}
\texttt{
“Can the following question-answer pair be restructured into a classification problem? Question: ‘{Question}’ Answer: ‘{Answer}’. If yes, respond with ‘yes’. If no, respond with ‘no’.”
}
\end{quote}
The GPT-4o mini model responded with “yes” or “no” based on whether the question-answer pair could be structured into a classification task.

\end{document}